\documentclass[journal,twoside,web]{ieeecolor}
\usepackage{tmi}
\usepackage{cite}
\usepackage[T1]{fontenc}
\usepackage{amsmath,amssymb,amsfonts}
\usepackage{multirow}
\usepackage{url}
\usepackage{algorithmic}
\usepackage{graphicx}
\usepackage{textcomp}
\def\BibTeX{{\rm B\kern-.05em{\sc i\kern-.025em b}\kern-.08em
    T\kern-.1667em\lower.7ex\hbox{E}\kern-.125emX}}
\begin{document}
\title{LMS-Net: A Learned Mumford-Shah Network For Few-Shot Medical Image Segmentation}
\author{Shengdong Zhang, Fan Jia, Xiang Li, Hao Zhang, Jun Shi, \IEEEmembership{Member, IEEE},\\ Liyan Ma, and Shihui Ying, \IEEEmembership{Member, IEEE}
\thanks{This work was supported by the National Key R \& D Program of China under Grant 2021YFA1003004.}
\thanks{Corresponding authors: Liyan Ma and Shihui Ying.  (e-mail: 
 liyanma@shu.edu.cn; shying@shu.edu.cn)}
\thanks{Shengdong Zhang and Hao Zhang are with the Department of Mathematics, School of Science, Shanghai University, Shanghai 200444, China. (e-mail:
zsd2@shu.edu.cn; Zhanghao123@shu.edu.cn).}
\thanks{Fan Jia is with the Department of Mathematics and Scientific Computing and Imaging (SCI) Institute, University of Utah, Salt Lake City, UT 84102, USA. (e-mail: fan.jia@utah.edu).}
\thanks{Xiang Li is with the school of Computer Science and Technology, East China Normal University, Shanghai, China. (e-mail:
51265901126@stu.ecnu.edu.cn).}
\thanks{Jun Shi is with the School of Communication and Information Engineering, Shanghai University, Shanghai 200444, China. (e-mail: junshi@shu.edu.cn).}
\thanks{Liyan Ma is with the School of Computer Engineering and Science,
Shanghai University, Shanghai 200444, China, and also with the School
of Mechatronic Engineering and Automation, Shanghai Key Laboratory
of Intelligent Manufacturing and Robotics, Shanghai University, Shanghai 200444, China. (e-mail: liyanma@shu.edu.cn).}
\thanks{Shihui Ying is with Shanghai Institute of Applied Mathematics and Mechanics and the School of Mechanics and Engineering Science, Shanghai University, Shanghai 200072, China. (e-mail: shying@shu.edu.cn).}}

\maketitle

\begin{abstract}
Few-shot semantic segmentation (FSS) methods have shown great promise in handling data-scarce scenarios, particularly in medical image segmentation tasks. However, most existing FSS architectures lack sufficient interpretability and fail to fully incorporate the underlying physical structures of semantic regions. To address these issues, in this paper, we propose a novel deep unfolding network, called the Learned Mumford-Shah Network (LMS-Net), for the FSS task. Specifically, motivated by the effectiveness of pixel-to-prototype comparison in prototypical FSS methods and the capability of deep priors to model complex spatial structures, we leverage our learned Mumford-Shah model (LMS model) as a mathematical foundation to integrate these insights into a unified framework. By reformulating the LMS model into prototype update and mask update tasks, we propose an alternating optimization algorithm to solve it efficiently. Further, the iterative steps of this algorithm are unfolded into corresponding network modules, resulting in LMS-Net with clear interpretability. Comprehensive experiments on three publicly available medical segmentation datasets verify the effectiveness of our method, demonstrating superior accuracy and robustness in handling complex structures and adapting to challenging segmentation scenarios. These results highlight the potential of LMS-Net to advance FSS in medical imaging applications. Our code will be available at: \url{https://github.com/SDZhang01/LMSNet}
\end{abstract}


\section{Introduction}
\label{sec:introduction}
\IEEEPARstart{I}{mage} segmentation is a fundamental task with numerous applications in clinical procedures, including disease diagnosis \cite{masood2015survey}, treatment planning \cite{chen2021deep,el2007} and treatment delivery. Fully supervised deep learning methods \cite{badrinarayanan2017segnet,chen2021transunet,ronneberger2015u} have shown impressive performance in segmentation tasks when trained on large, well-annotated datasets. However, their applicability in medical settings is limited by the scarcity of annotated data for specific classes, which poses a significant challenge in real-world scenarios. This challenge motivates the development of few-shot semantic segmentation (FSS) methods, which aim to segment images using only a few labeled examples, a scenario particularly relevant in medical imaging tasks.

\begin{figure*}
    \centering
    \includegraphics[width=1\linewidth]{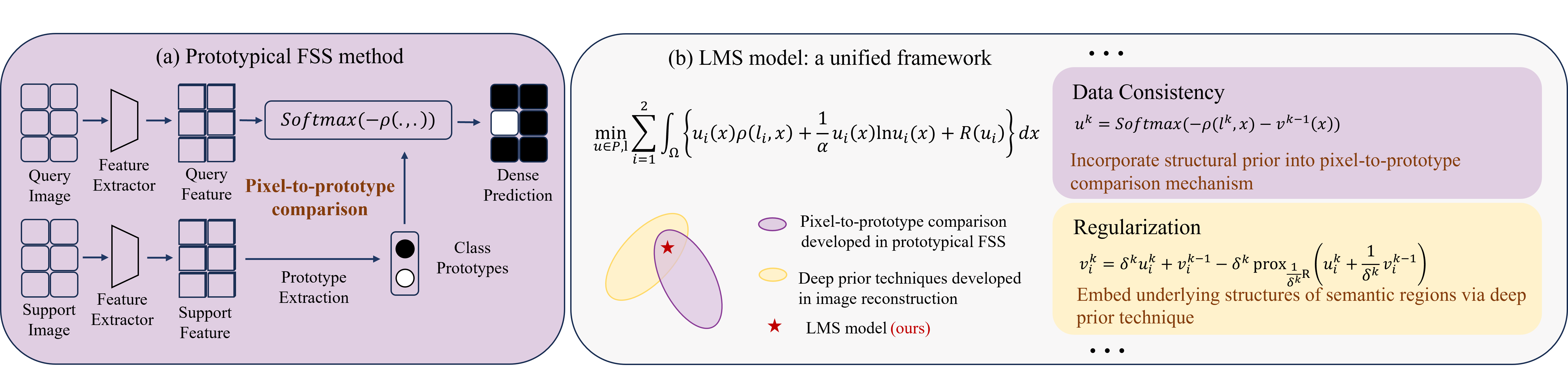}
    \caption{Illustration of the pixel-to-prototype comparison mechanism in prototypical FSS methods and how our LMS model integrates pixel-to-prototype comparison and deep prior techniques into a unified framework.}
    \label{fig:overview}
\end{figure*}

Recently, prototypical FSS networks \cite{yu2021location,Ref26,huang2023rethinking} have emerged as a promising alternative, enabling the segmentation of novel classes using only a limited number of labeled examples. These models leverage deep networks to perform pixel-to-prototype comparison in a latent space, guided by prototypes extracted from labeled samples, as shown in Fig. \ref{fig:overview}(a). Despite their success, they still have two evident limitations: (1) the intrinsic prior knowledge underlying the FSS task, such as the structural prior of the semantic regions (e.g., spatial continuity), is not fully utilized; (2) the overall framework lacks sufficient interpretability, which is a topic of most concern for medical AI.  

In contrast, variational methods\cite{chan2001active,vese2002multiphase,Ref3}, as traditional approaches to image segmentation, offer strong interpretability. Among these methods, the piecewise constant Mumford-Shah model \cite{Ref3} is particularly notable. This approach strikes a balance between a data fidelity term, which encourages each pixel to be assigned to the prototype with minimal cost, and a regularization term (i.e. total variation) that captures the structural prior of the segmented regions. However, its reliance on handcrafted priors, sensitivity to intensity variance within the same semantic region, and the need for manual initialization for each image limit its adaptability to complex medical images, compared to learning-based approaches. 

To address these limitations, we should integrate the merits of variational methods and deep neural networks by establishing the relationship between them. Specifically, the data fidelity term of the Mumford-Shah model aligns naturally with the pixel-to-prototype comparison mechanism in prototypical FSS methods. Meanwhile, prototypical FSS networks, with their strong feature representation capabilities, facilitate straightforward prototype initialization within the variational framework. Furthermore, this integration paves the way for adopting deep prior techniques within the realm of the FSS task. Originally developed for image reconstruction, deep prior techniques \cite{Ref5,Ref13,Ref18} have achieved notable success by capturing spatial structures more effectively than traditional priors. However, the distinct physical mechanisms underlying segmentation differ significantly from those of reconstruction, making it challenging to introduce deep prior techniques into FSS.

A key observation is that, with an appropriate splitting method, the prior term in the Mumford-Shah model can be decoupled from the data fidelity term, which corresponds to FSS, and reformulated as a denoising problem. This allows a CNN-based denoiser to better capture intricate anatomical structures. As a result, the deep prior technique can also be naturally incorporated into the FSS framework under the Mumford-Shah model.

Based on this observation, we propose the Learned Mumford-Shah Network (LMS-Net) to tackle the FSS task. Compared with other empirically designed networks, the proposed LMS-Net is derived from a learned Mumford-Shah type variational model, referred to as the LMS model, which naturally integrates the insights of pixel-to-prototype comparison and deep prior techniques, as shown in Fig .\ref{fig:overview}. The LMS model is divided into two primary tasks: prototype update, which is achieved through a momentum-based approach, and mask update, which uses a primal-dual algorithm. These two tasks are alternately optimized to effectively solve the model. Then, the iterative steps of the proposed algorithm are unfolded into distinct network modules, resulting in  LMS-Net, a coherent and interpretable end-to-end deep unfolding network. The main contributions of this paper are as follows:

\begin{itemize}[]
    \item We propose a novel learned Mumford-Shah model to tackle the FSS task. This model describes the pixel-to-prototype mechanism through the data fidelity term and incorporates the underlying physical structures of semantic regions using a deep prior.
    \item  Building on the LMS model, we employ a momentum-based approach for prototype update and a primal-dual algorithm for mask update, enabling alternate optimization to effectively solve the model.
    \item We introduce the Momentum Update Transformer (MUT) to incorporate information from previous stages for prototype updates, and design the PD-Net, which adopts a primal-dual update structure for mask refinement. Therefore, the entire optimization process is formulated as a deep unfolding network, which allows for end-to-end learning with clear interpretability.
    \item To validate the superiority and robustness of the proposed method, we perform extensive experiments on three widely used medical image datasets. The results, supported by detailed visualizations and ablation studies, further underscore the advantages of our approach.
\end{itemize}

The remainder of this paper is organized as follows. In Section \ref{sec:1}, we review the related work on variational segmentation methods, few-shot semantic segmentation methods, and deep prior techniques. In Section \ref{sec:8}, we introduce the learned Mumford-Shah model to tackle the FSS task along with the corresponding optimization algorithm. Based on this model, the proposed Learned Mumford-Shah Network is introduced in Section \ref{Sec: 9}. Finally, Section \ref{sec:2} shows the numerical results.

\section{ Related Work}
\label{sec:1}
\subsection{Variational Image Segmentation}
Image Segmentation has been traditionally treated as an energy minimization problem which imposes a tradeoff between a data fidelity term and a regularization term.  The Mumford-Shah (MS) model \cite{Ref3} is one of the most well-known variational models.


Specifically, the two-phase piecewise constant model is formulated as follows: given an image $I$, find indicator functions $u_1$, $u_2$ corresponding to the partition $\{\Omega_1, \Omega_2\}$ of the image domain $\Omega$ and mean values $l_1,l_2$ such that the following functional is minimized:

\begin{equation}
\label{deqn_ex5a}
\begin{aligned}
\min _{u\in P,l} & \sum_{i=1}^2 \int_{\Omega}\left\{u_i(x) \rho\left(l_i, x\right)+\lambda\left|\nabla u_i\right|\right\} d x, \\ 
\end{aligned}
\end{equation}
where the set $P$ is defined as:
\begin{equation}
\begin{aligned}
P=\left\{u: \Omega \mapsto \mathbb{R}^2 \mid u(x) \in \Delta_{+}, \forall x \in \Omega\right\} ,
\end{aligned}
\end{equation}
and the simplex constraint $\Delta_+$ is:
\begin{equation}
\begin{aligned}
\Delta_{+}=\left\{u \in \mathbb{R}^2 \mid \sum_{i=1}^2 u_i=1 ; u_i \geq 0, i=1,2\right\},
\end{aligned}
\end{equation}
 $\lambda$ is the balancing coefficient and $\rho\left(l_i, x\right)$ estimates the pixel-wise error of assigning $l_i$ to the pixel $x$. For example, $\rho\left(l_i, x\right)$ can be defined as $\left|I(x)-l_i\right|^2$, which measures the squared difference between the pixel intensity $I(x)$ and the mean value $l_i$.

Many algorithms have been developed to solve this relaxed problem \cite{Ref10,Ref2,Ref17}. These methods have proved to be very efficient for segmenting piecewise constant types of cartoon-like images, but they often struggle with the intricate structures and intensity variations present in complex segmentation tasks. 

Recently, neural networks have been integrated into some variational segmentation approaches \cite{Ref7,Ref8,Ref9}. Jia et al. proposed an energy function with a learnable data term and a handcrafted prior term (i.e. total variation) \cite{Ref7}. The solution derived from this learned energy model is differentiable with respect to the parameters of the energy function, which enables the iterative optimization process to be interpreted as a cascaded network. Liu et al. further investigated the incorporation of various handcrafted priors into neural network architectures\cite{Ref8}. By leveraging the powerful representational capacity of neural networks, these methods exhibit improved performance when handling complex images. However, these methods still require hundreds of iterations to converge, resulting in high computational costs, and the predefined priors may not be well-suited for diverse image characteristics.

Our method can also be derived from a learned energy function, which is closely related to \eqref{deqn_ex5a}. In contrast to previous works, we utilize the specific few-shot segmentation setting to derive a favorable initial mean value $l_i$ in the data fidelity term, enhancing class-level feature representation. Additionally, we replace traditional total variation (TV) regularization with the deep denoising prior, drawing inspiration from deep unfolding methods used in image reconstruction tasks. This reformulation leverages the data-driven nature of deep priors, enabling the model to adaptively capture complex structures and patterns in semantic regions.

\subsection{Few-Shot Semantic Segmentation}

In contrast to variational models operating under a data-free paradigm or supervised learning methods that require large amounts of labeled data for a specific class, the few-shot segmentation task is introduced to segment unseen classes using a limited amount of labeled data as support. Dong et al. proposed a prototypical episode segmentation paradigm \cite{Ref24}, where a global prototype for each class is derived from the support set, followed by per-pixel classification via prototype comparison. This paradigm has since been widely adopted and extended in numerous studies \cite{Ref25,Ref26,Ref27,Ref28,Ref29}. Tang et al. introduced a prototype network with recurrent mask refinement \cite{Ref29}, where the previous query prediction iteratively refines the query features. Zhu et al. and Liu et al. proposed transformer-like network architectures for adaptively updating prototypes\cite{Ref25, Ref27}.

Despite their effectiveness, these deep learning methods often function as black boxes, directly learning mappings from images to probability masks without offering interpretability. In contrast, our LMS-Net is derived from a variational model, where each component of the network is designed with clear interpretability. 
\subsection{Deep Prior Technique}
Many imaging problems can be formulated as variational problems based on total variation. Inspired by the success of deep learning, image reconstruction has embraced deep unfolding methods, which primarily focus on learning the prior term using neural networks. Some approaches \cite{Ref4,Ref5,Ref18,Ref19} attempted to generalize the handcrafted transforms in the prior term into learnable transforms. ADMM-CSNet \cite{Ref4} utilizes several learnable linear filters for compressed sensing (CS) magnetic resonance imaging. ISTA-Net \cite{Ref5} further adopts nonlinear transforms and performs well in general CS tasks. These networks correspond strictly to traditional algorithms of an explicit energy function.

An alternative strategy is to implicitly model the prior term by solving it as a denoising subproblem using a CNN. These approaches are driven by the fact that the proximal operator of the prior term can be interpreted as a denoising subproblem \cite{heide2014flexisp}. For instance, Adler et al. proposed a network derived from the Primal Dual Hybrid Gradient (PDHG) method \cite{Ref2} and replaced the proximal operator with a shallow CNN \cite{Ref1}. This approach has demonstrated superiority in solving various inverse problems \cite{Ref13,Ref14,Ref16}. 

Building on these ideas, our approach utilizes and unfolds the primal-dual algorithm \cite{Ref10} for image segmentation. Unlike previous works, which focus on image reconstruction, the proposed LMS-Net uniquely integrates the deep denoiser technique into segmentation tasks, marking a novel contribution in this domain.
\section{METHODS}
\subsection{Learned Mumford-Shah Model}
\label{sec:8}

\subsubsection{Problem Definition}
\label{sec:3}
The goal of few-shot semantic segmentation is to obtain a model that can segment unseen semantic classes with only a few labeled images of this unseen class without retraining. In FSS, the entire dataset is typically divided into two distinct subsets. One is the training set $\mathcal{D}_{t r}$ with training semantic classes $\mathcal{C}_{t r}$, the other is the testing set $\mathcal{D}_{t e}$ with unseen testing classes $\mathcal{C}_{t e}$, where $\mathcal{C}_{t r}\cap \mathcal{C}_{t e}=\emptyset$. The mainstream setting adopts
the episodic training and testing scheme. Each episode consists of a support set $S$ and a query set $Q$ for a specific class $c$. $S=\left\{\left(I_t^s, u_{c,t}^s\right) \mid c=\right.1, \ldots, N ; t=1, \ldots, T\}$ contains $T$ images $I^s$ with $N$ classes and the corresponding binary masks $u_c^s$, while $Q=\{I^q\}$ only contains
images $I^q$ to be segmented. An episode consisting of a support-query pair $(S,Q)$ is called the N-way T-shot sub-task. For simplicity, we focus on 1-way 1-shot learning, which is adopted in most works \cite{Ref28,Ref26}.

\label{sec:5}
\subsubsection{Model Formulation}

Motivated by the effectiveness of pixel-to-prototype comparison, a core concept in prototypical FSS, and the capability of deep priors to capture complex spatial structures, we leverage the Mumford-Shah model as a natural mathematical foundation to integrate these insights into a unified framework. Besides, entropy regularization is incorporated to ensure the smoothness of the solution. Based on this, we formulate our LMS model by transforming the classical MS model \eqref{deqn_ex5a} into the following optimization problem:
\begin{equation}
\label{deqn_ex11a}
\begin{aligned}
\min _{u\in P, l} & \sum_{i=1}^2 \int_{\Omega}\left\{u_i(x) \rho\left(l_i, x\right)+\frac{1}{\alpha} u_i(x) \ln u_i(x) + R(u_i)\right\} d x,  
\end{aligned}
\end{equation}
where $R(u)$ denotes the deep prior term, $\rho$ is given by
\begin{equation}
\label{deqn_ex9a}
\begin{aligned}
\rho(l_i,x) = -\frac{F^q(x)\cdot l_i}{\|F^q(x)\|\|l_i\|},
\end{aligned}
\end{equation}
where $\cdot$ refers to the inner product, and \eqref{deqn_ex9a} is a variant of cosine distance between $F^q(x)$ and $l_i$. $F^q\in \mathbb{R}^{H \times W \times C}$ are the latent features for the query image $I^q$, generated by the feature extractor $f_\theta(.)$. $l_i\in \mathbb{R}^{C},i =1,2$ denote the prototypes corresponding to the foreground and background, respectively. By default, we assume that $i=1$ represents the foreground and $i=2$ represents the background. The second term of \eqref{deqn_ex11a}, referred to as entropy regularization, facilitates soft thresholding in the segmentation process. This property is advantageous for backpropagation within the network. The proposed LMS model \eqref{deqn_ex11a} differs from \eqref{deqn_ex5a} in two main aspects:
\begin{itemize}
    \item Instead of estimating the pixel-wise error in the intensity space, we perform this estimation in the latent space. This term describes the pixel-to-prototype comparison mechanism corresponding to FSS.
    \item TV prior in \eqref{deqn_ex5a}  is extended to a deep prior in \eqref{deqn_ex11a}, which can be adaptively learned from data. This term is designed to preserve the underlying physical structures of semantic regions.
\end{itemize}
\subsubsection{Optimization Algorithm}
To solve this problem, we alternately update each variable with other variables fixed. It leads to the following two subproblems:
\begin{equation}
\label{deqn_ex12a}
\begin{aligned}
\min _{l} & \sum_{i=1}^2 \int_{\Omega}u_i(x) \rho\left(l_i, x\right) d x, \\ 
\end{aligned}
\end{equation}
\begin{equation}
\label{deqn_ex13a}
\begin{aligned}
\min _{u\in P} & \sum_{i=1}^2 \int_{\Omega}\left\{u_i(x) \rho\left(l_i, x\right)+\frac{1}{\alpha} u_i(x) \ln u_i(x) +  R(u_i)\right\} d x. \\ 
\end{aligned}
\end{equation}

\textbf{updating $l$}: Since scaling $l$ does not affect the solution of this problem, the solution for $l$ can be expressed as:
\begin{equation}
\label{deqn_ex20a}
\begin{aligned}
l^k_i=\beta_i\frac{\int_{\Omega} F^q(x) \odot u^{k-1}_{i}(x)dx}{\int_{\Omega} u^{k-1}_i(x)dx}, \beta_i>0,
\end{aligned}
\end{equation}
where $\odot$ denotes the Hadamard product, $\beta_i$ is a scaling factor, and by default, we set $\beta_i=1 $ for $i=1,2$. \eqref{deqn_ex20a} can be interpreted as performing Masked Average Pooling (MAP) on the query features, for both the foreground and background. However, this operator may cause a significant loss of prototype information from previous stages. A momentum update approach will be proposed in the next section to effectively incorporate information from previous prototypes into the current stage.

\textbf{updating $u$}:
It is well known that this problem corresponds to the convex relaxed Potts model \cite{Ref20}. Since the primal-dual algorithm has been proven effective for this problem \cite{Ref2,Ref10}, We reformulate \eqref{deqn_ex13a} as a saddle-point problem, expressed as follows:
\begin{equation}
\label{deqn_ex14a}
\begin{aligned}
\min _{u \in P} \max _{v} E(u, v):= 
&\sum_{i=1}^2 \int_{\Omega} \bigg\{ u_i(x) \big(\rho\left(l^k_i, x\right)+v_i(x)\big)  \\
&\quad  +\frac{1}{\alpha} u_i(x) \ln u_i(x) - R^*(v_i) \bigg\} dx,
\end{aligned}
\end{equation}
where $R^*$ denotes the Fenchel conjugate of $R$, $v$ is the dual variable. By alternately iterating and optimizing the primal and dual
problems. The optimization for the $k^{th}$ iteration can be represented as:
\begin{equation}
\label{deqn_ex15a}
\begin{aligned}
u^k=\operatorname*{arg\,min}_{u \in P} & \sum_{i=1}^2 \int_{\Omega}\bigg\{u_i(x) (\rho\left(l^k_i, x\right)+v^{k-1}_i(x))\\
& +\frac{1}{\alpha} u_i(x) \ln u_i(x)\bigg\} d x, 
\end{aligned}
\end{equation}

\begin{equation}
\label{deqn_ex16a}
\begin{aligned}
v^k = \operatorname*{arg\,max}_{v } 
\sum_{i=1}^2 \int_{\Omega} 
\bigg\{ 
- \frac{1}{2\delta^k}
\Big(
v_i(x) - (v_i^{k-1}(x) \\
+ \delta^k u_i^k(x))
\Big)^2 
- R^*(v_i)
\bigg\} d x.  
\end{aligned}
\end{equation}
The data fidelity term and the deep prior term are decoupled into \eqref{deqn_ex15a} and \eqref{deqn_ex16a}, respectively. \eqref{deqn_ex15a} consists of a linear term and an entropy regularization term, and the closed-form solution is given by:
\begin{equation}
\label{deqn_ex17a}
\begin{aligned}
u^k=\operatorname*{Softmax}(\alpha(-\rho\left(l^k, x\right)-v^{k-1}(x))).
\end{aligned}
\end{equation}
The intermediately predicted mask $u^k$ depends on both pixel-wise error and a fixed estimate of $v$. \eqref{deqn_ex17a}  can be viewed as a generalization of the pixel-to-prototype comparison mechanism in prototypical FSS methods, where $v$, as a correction to the physical structures, is additionally considered. This simple operator can be easily incorporated into a network.

The $v$-subproblem is also called the proximity operator of $\delta^kR^*(v_i)$. Based on the Moreau decomposition \cite{Ref21,Ref22}, this problem is equivalent to
\begin{equation}
\label{deqn_ex18a}
\begin{aligned}
v_i^k=\delta^ku_i^k + v_i^{k-1} - \delta^k\operatorname*{prox_{\frac{1}{\delta^k}R}}(u_i^k + \frac{1}{\delta^k}v_i^{k-1}),
\end{aligned}
\end{equation}
where the proximal mapping of the regularizer $\frac{1}{\delta^k}R$, denoted by $\operatorname*{prox_{\frac{1}{\delta^k}R}}(u_i^k + \frac{1}{\delta^k}v_i^{k-1})$, is defined as:
\begin{equation}
\label{deqn_ex19a}
\begin{aligned}
&\operatorname{prox}_{\frac{1}{\delta^k}R}
\big(u_i^k + \frac{1}{\delta^k}v_i^{k-1}\big) = 
\arg \min _{v_i} \int_{\Omega} 
\bigg\{
\frac{\delta^k}{2} 
\Big(
v_i(x) \\
&- \big(\frac{1}{\delta^k}v_i^{k-1}(x) 
+ u_i^k(x)\big)
\Big)^2 
+ R(v_i)
\bigg\} dx.
\end{aligned}
\end{equation}
\begin{figure*}
    \centering
    \includegraphics[width=1\linewidth]{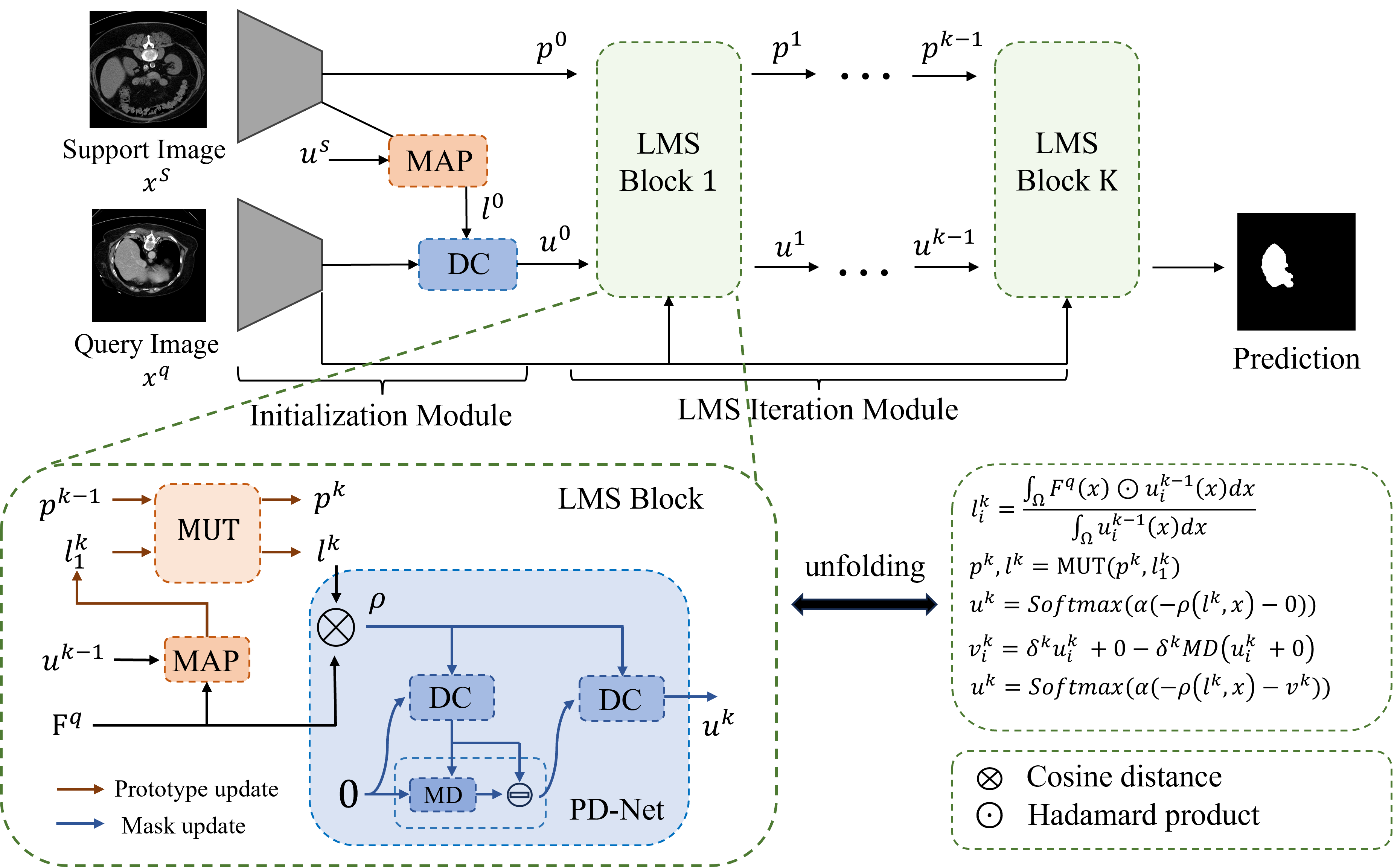}
    \caption{The overall structure of the proposed Learned Mumford-Shah Network (LMS-Net)  for the FSS task.}
    \label{fig:1}
\end{figure*}
From a Bayesian perspective, this problem corresponds to Gaussian denoising on $u_i^k + \frac{1}{\delta^k}v_i^{k-1}$. Thus, $v_i^k$ in \eqref{deqn_ex18a} represents the noise component in $u_i^k + \frac{1}{\delta^k}v_i^{k-1}$, which embeds the underlying physical structures of semantic regions. As demonstrated, under an appropriate splitting algorithm, \eqref{deqn_ex13a} can be separated into a simple data subproblem associated with the FSS mechanism and a prior subproblem, which corresponds to image denoising. Their solutions correspond to \eqref{deqn_ex17a} and \eqref{deqn_ex18a}, respectively. In fact, modern convex optimization algorithms for solving various image reconstruction problems can also be separated into distinct data subproblems and identical prior subproblems. Significant success has been achieved in solving the denoising subproblem using powerful CNN-based denoisers. We adopt this idea to unfold our optimization algorithm into a network architecture, with a more detailed discussion provided in the next section.

\subsection{Learned Mumford-Shah Network}
\label{Sec: 9}
\subsubsection{Network Overview}

Fig. \ref{fig:1} shows the framework of our LMS network for the FSS task, which consists of the Initialization Module and the LMS Iteration Module. The Initialization Module initializes the prototypes $l^0$ and the mask $u^0$ in the context of the FSS setting. Additionally, multiple representative prototypes $p^0$ from the support image are initialized to preserve prototype information across iterations. The LMS Iteration Module consists of $K$ LMS Blocks, which represent $K$ iterations of the algorithm designed to solve \eqref{deqn_ex11a}. Specifically, each LMS block comprises three key components: a MAP operator, MUT, and PD-Net.

The MAP operator, corresponding to the closed-form solution in \eqref{deqn_ex20a}, updates the prototype $l$ based on the current predicted mask. To reduce the information loss of prototypes in previous stages, Momentum Update Transformer (MUT) is proposed to establish a deep and flexible long-term information path across all iterations. 

\begin{figure*}
    \centering
    \includegraphics[width=1\linewidth]{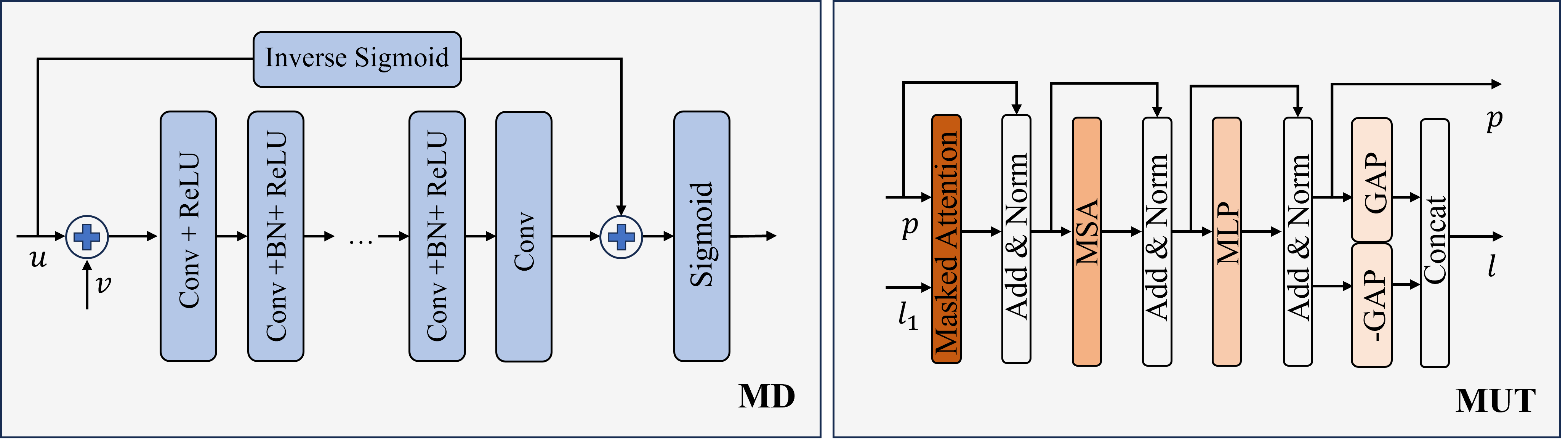}
    \caption{The structure of the proposed Mask Denoiser and Momentum Update Transformer.}
    \label{model_structure}
\end{figure*}

In PD-Net, $u$ and $v$ are iteratively updated to refine the predicted mask. First, the Data Consistency layer (DC), derived from the closed-form solution in \eqref{deqn_ex17a}, updates the mask $u$ based on the current prototype $l$. Next, the proximal mapping in \eqref{deqn_ex18a} is treated as a denoising process, and a shallow CNN denoiser, referred to as Mask Denoiser (MD), is introduced to learn the deep prior. The updated $v$ is obtained by applying a skip connection to the denoiser, following \eqref{deqn_ex18a}. Finally, DC is applied again to refine the mask $u$ based on \eqref{deqn_ex17a}. Detailed explanations of each module will be provided in subsequent sections.
\subsubsection{Initialization Module}
We first adopt a shared weight backbone $f_\theta(.)$  to generate support and query latent features, i.e. $F^s\in \mathbb{R}^{H \times W \times C}$ and $F^q\in \mathbb{R}^{H \times W \times C}$, for the support image $I^s$ and the query image $I^q$, respectively. A good initial guess is essential for classical variational segmentation models, but in FSS, it can be easily obtained. Specifically, considering  \eqref{deqn_ex20a} and the fact that the support image has a ground truth mask, the optimal prototypes for the support image are given by:
\begin{equation}
\label{deqn_ex21a}
\begin{aligned}
l^s_i=\frac{\int_{\Omega} F^s(x) \odot u^{s}_{i}(x)dx}{\int_{\Omega} u^{s}_{i}(x)dx}.
\end{aligned}
\end{equation}
Given that the foreground of the support and query images belong to the same class, we set $l_1^0=l^s_1$, $l_2^0=-l^s_1$. The mask $u^0$ is initialized following \eqref{deqn_ex17a} with $v^0=0$:
\begin{equation}
\label{deqn_ex22a}
\begin{aligned}
u^0=\operatorname*{Softmax}(\alpha(-\rho\left(l^0, x\right))).
\end{aligned}
\end{equation}
Moreover, multiple representative prototypes $p^0\in \mathbb{R}^{N_p \times C} $ are generated to preserve the original prototypical information. Specifically, the foreground mask of the support image $u_1^s$ is further divided into $N_p$ regions using a superpixel algorithm based on Voronoi partitioning. $p^0$ can then be obtained by applying the MAP operator to each of the $N_p$ regions individually.


\subsubsection{LMS Iteration Module}
\label{Iter}
As shown in Fig. \ref{fig:1}, the LMS Iteration Module  consists of $K$ LMS blocks, representing $K$ iterations of the algorithm for solving \eqref{deqn_ex11a}. Specifically, each LMS block includes a MAP operator and MUT for solving the prototype update task \eqref{deqn_ex12a}, as well as PD-Net for solving the mask update task \eqref{deqn_ex13a}.

\textbf{MAP:} In MAP, the prototype is computed to effectively represent the pixel features within a region and can be directly formulated as in \eqref{deqn_ex20a}. A closed-form solution is then obtained given $(u^{k-1}, F^q)$.

\textbf{MUT:} To explicitly model the long-range dependencies of prototypes across all unfolded stages, we employ a transformer layer, which has demonstrated both stability and effectiveness in balancing historical and current states.

The MUT layer follows the MAP layer and updates both the query prototype $l$ and the set of multiple representative prototypes $p$, as illustrated in Fig. \ref{model_structure}. This update process is formulated as:
\begin{equation}
\label{deqn_ex27a}
\begin{aligned}
p,l = \operatorname*{MUT}(p,l_1),  
\end{aligned}
\end{equation}
where the foreground prototype $l_1$ serves as the input to MUT. The updated query prototype $l$ integrates information from the prototypes in the previous stage and is subsequently passed to PD-Net. Simultaneously, the representative prototypes $p$ are refined by incorporating feedback from $l$, effectively maintaining a dynamic memory across stages.

The architecture of MUT is explicitly defined as:
\begin{equation}
\label{deqn_ex31b}
\begin{aligned}
p =& LN(\operatorname*{Softmax}(\mathcal{M}+pl_1^T)l_1+p), \\
p =& LN(MSA(p)+p), \\
p =& LN(MLP(p)+p), \\
l =& [GAP(p), -GAP(p)],
\end{aligned}
\end{equation}
where $LN(.)$ denotes the layer normalization operator, $MSA(.)$ denotes the Multihead Self-Attention, $MLP(.)$ denotes the Multi-Layer Perceptron and $GAP(.)$ denotes the global average pooling. The masking matrix $\mathcal{M}\in \mathbb{R}^{N_p \times 1} $ is defined by:
\begin{equation}
\label{deqn_ex30b}
\mathcal{M}(n)= \begin{cases}0, & \text { if } p_nl_1^T>\epsilon \\ -\infty, & \text { otherwise }\end{cases}, n=1,...,N_p
\end{equation}
where $\epsilon = (min(pl_1^T)+mean(pl_1^T))/2$. In this design, $\mathcal{M}$ ensures that only relevant prototypes contribute to the updates, based on their similarity with the query prototype $l_1$. The $\text{MSA}$ mechanism and $\text{MLP}$ further enhance the flexibility and capacity of information propagation, while $\text{GAP}$ aggregates representative prototypes to form the updated query prototype $l$.

Thus, the proposed MUT layer effectively enhances information transmission across all stages, addressing potential loss of long-range dependencies.

\textbf{PD-Net:} PD-Net is derived from the iterative procedures of primal dual algorithm for optimizing \eqref{deqn_ex13a}. In this module, we take the query latent feature $F^q$ and the previous prototype $l^{k}$ as input, and we output the updated mask $u^{k}$. The iteration blocks are unfolded following the updating rules of \eqref{deqn_ex17a}, \eqref{deqn_ex18a}. In each inner loop, the dual variable $v$ is initialized to zero. Then the primal variable $u^k$ is initialized with DC corresponds to \eqref{deqn_ex17a}:
\begin{equation}
\label{deqn_ex23a}
\begin{aligned}
u^k=\operatorname*{Softmax}(-\alpha(\rho\left(l^{k}, x\right))).
\end{aligned}
\end{equation}
The key issue of unfolding the algorithm of \eqref{deqn_ex18a} is how to represent the proximal operator $prox_R(.)$. Motivated by the denoising interpretation of the proximal operator and the powerful performance of CNNs in image denoising\cite{Ref23}, we replace $prox_R(.)$ with a shallow CNN. The refined mask $u^{k+1}$ can be obtained by iterative application of DC. 

Thus, at the $k^{th}$ stage, PD-Net is built by:
\begin{equation}
\label{deqn_ex25a}
\left\{\begin{array}{l}
\displaystyle u^k=\operatorname*{Softmax}(\alpha(-\rho\left(l^k, x\right)-0)), \\
v_i^k=\delta^ku_i^k + 0 - \delta^k\operatorname*{MD}(u_i^k +0),\\
u^k=\operatorname*{Softmax}(\alpha(-\rho\left(l^k, x\right)-v^{k})),
\end{array}\right.
\end{equation}
where MD denotes the mask denoiser built on DnCNN \cite{Ref23} with 5 convolution layers. The structure of MD is illustrated in Fig. \ref{model_structure}. Compared to DnCNN, we make two slight modifications to better adapt it for mask denoising. First, we add a Sigmoid operator in the final layer to constrain the output within the range of $0$ to $1$. Second, an inverse Sigmoid transformation is applied to $u$ and incorporated before the Sigmoid layer.  
This modification is introduced to improve the stability of the network training.

It should be noted that if we do not consider the update of the prototype $l$ in \eqref{deqn_ex11a} and directly input $l^0$ into PD-Net, this network can be regarded as the unfolded network for the Potts model \cite{Ref20}. In this case, the resulting network, referred to as fLMS-Net (LMS-Net with fixed prototypes), is a simplified version of LMS-Net. This network can also be utilized to address the FSS task, and its effectiveness will be demonstrated in Section \ref{sec:2}.
\subsection{Loss Function}
Our LMS-Net is trained episodically in an end-to-end manner. Given the ground truth mask
 $\hat{u}^q$ and the predicted mask $u$ for the query image $I^q$, we can define the cross-entropy loss between $\hat{u}^q$ and $u$ as our main loss:
\begin{equation}
\label{deqn_ex26a}
\begin{aligned}
L_{ce} =  \sum_{i=1}^2 \frac{\int_{\Omega} \hat{u}^q_i(x) \ln u_i(x) d x}{\int_{\Omega} 1 d x}.  
\end{aligned}
\end{equation}

Additionally, we introduce a prototype alignment regularization loss $L_{par}$ to encourage consistency between the support and query prototypes, following common practice \cite{Ref26,Ref28}. As a result, we obtain the overall loss function:
\begin{equation}
\label{deqn_ex29a}
\begin{aligned}
L =  L_{ce} + L_{par}.
\end{aligned}
\end{equation}
\section{Experiments}
\label{sec:2}

\begin{table*}[ht!]
\caption{PERFORMANCE COMPARISON (IN DSC $\%$) OF LMS-Net AND EXISTING METHODS ON THREE MEDICAL DATASETS.}
\label{tab:1}
\centering
\resizebox{\textwidth}{!}{
\begin{tabular}{l c c c c c c c c c c c c c c c}
\hline
\multirow{2}{*}{Method} & \multirow{2}{*}{Prototype Update} & \multicolumn{5}{c}{Abd-CT} & \multicolumn{5}{c}{Abd-MRI} & \multicolumn{4}{c}{CMR} \\
\cline{3-7} \cline{8-12} \cline{13-16}
& & LK & RK & Liver & Spleen & Mean & LK & RK & Liver & Spleen & Mean & LV-BP & LV-MYO & RV & Mean \\
\hline
ADNet & no & 72.13 & 79.06 & 77.24 & 63.48 & 72.97 & 73.86 & 85.80 & 82.11 & 72.29 & 78.51 & 87.53 & 62.43 & 77.31 & 75.76 \\
fLMS-Net  & no & 72.91 & 68.15 & 79.93 & 74.87 & \textbf{73.97} & 78.28 & 89.02 & 82.98 & 72.23 & \textbf{80.63} & 89.72 & 69.39 & 80.59 & \textbf{79.90} \\
RPT & yes & 77.05 & 79.13 & 82.57 & 72.58 & 77.83 & 80.72 & 89.82 & 82.86 & 76.37 & 82.44 & 89.90 & 66.91 & 80.78 & 79.19 \\
LMS-Net & yes & 82.46 & 79.32 & 78.57 & 82.98 & \textbf{80.83} & 82.85 & 89.34 & 83.71 & 74.97 & \textbf{82.72} & 89.89 & 68.98 & 81.72 & \textbf{80.19} \\
\hline
\end{tabular}
}
\end{table*}

\begin{figure*}
    \centering
    \includegraphics[width=1\linewidth]{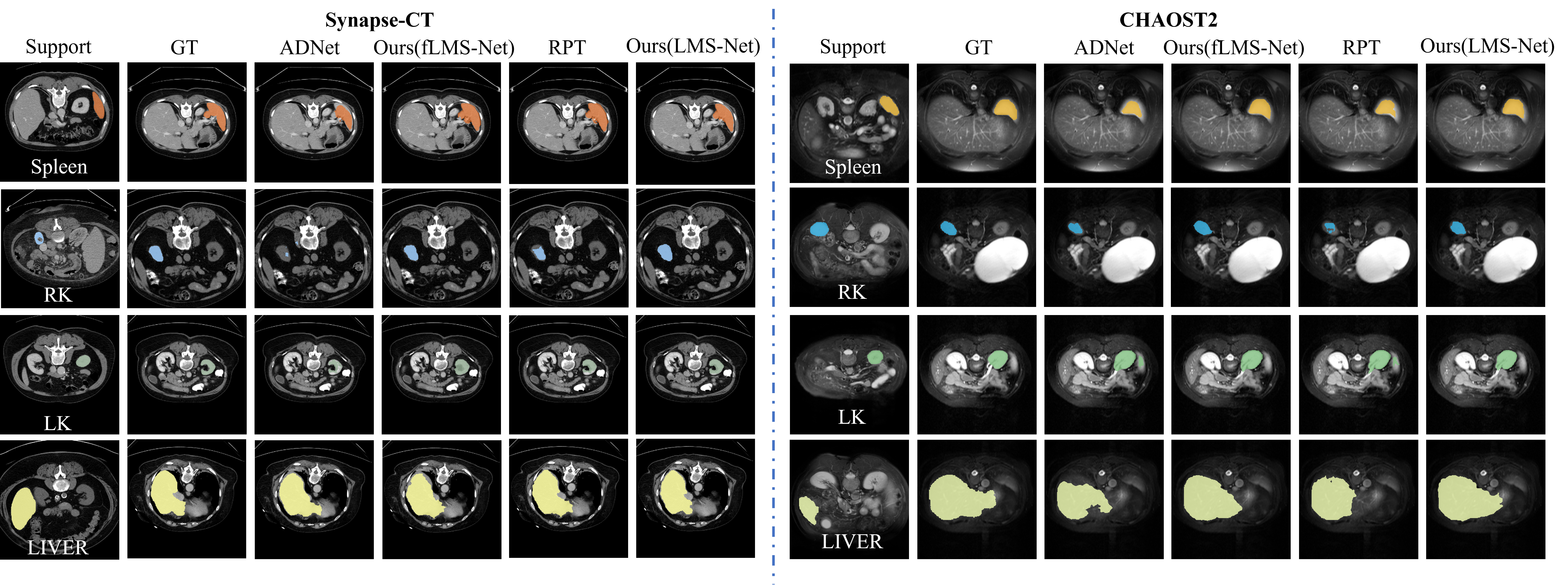}
    \caption{Qualitative comparison of our proposed LMS-Net with other medical FSS methods on the Synapse-CT and CHAOS-T2 datasets. GT is the ground truth.}
    \label{CT_MR}
\end{figure*}

\subsection{Datasets}
We evaluate the proposed method on three representative publicly available medical image datasets, including two abdominal organ datasets for MRI and CT (CHAOS-T2 and Synapse) and one cardiac dataset for MRI (MS-CMRSeg). Specifically,
\begin{itemize}[]
    \item \textbf{Synapse-CT}\cite{landman2015miccai} originates from the MICCAI 2015 Multi-Atlas Abdomen Labeling Challenge and includes $30$ 3D CT scans with a total of $3779$ axial slices. For evaluation, we focus on four specific organs: left kidney, right kidney, liver, and spleen.
    \item \textbf{CHAOS-MRI}\cite{kavur2021chaos} is part of the 2019 ISBI Combined Healthy Abdominal Organ Segmentation Challenge. It consists of $20$ 3D T2-SPIR abdominal MRI scans, each containing approximately $36$ slices. The same four organs as in the Synapse-CT dataset are selected: left kidney, right kidney, liver, and spleen.
    \item \textbf{CMR} \cite{zhuang2018multivariate}  is derived from the MICCAI 2019 Multi-Sequence Cardiac MRI Segmentation Challenge (bSSFP fold). It includes $35$ 3D cardiac MRI scans, each containing about $13$ slices, with three distinct cardiac labels: blood pool (LV-BP), left ventricle myocardium (LV-MYO), and right ventricle myocardium (RV).
\end{itemize}
\subsection{Baselines and Implementation Details}
To evaluate the effectiveness of our proposed method, we conducted a comparative analysis with two state-of-the-art approaches in the domain of few-shot medical image segmentation.

ADNet \cite{Ref26} is a representative method for prototypical few-shot segmentation. It extracts class prototypes from the support image and performs pixel-to-prototype comparison to segment the query image.  

RPT \cite{Ref25} builds upon the original ADNet by generating regional prototypes from the support image and leveraging a variant of the transformer to refine these prototypes.

Our network is implemented in the PyTorch framework and is trained end-to-end with a cascade stage $K$ set to $2$. All experiments are conducted on an NVIDIA RTX 3090 GPU with $24$ GB of memory.  Following the implementation details in \cite{Ref28,Ref26}, we adopt similar preprocessing techniques and a self-supervised training approach. To be specific, all 2D slices are extracted from 3D volumetric scans and resized to a spatial resolution of $256 \times 256$. We utilize ResNet-101 \cite{Ref32} as the backbone $f_\theta(.)$, which has been pretrained on a subset of the MS-COCO dataset \cite{Ref33}. When $256\times 256$ slices are passed through ResNet-101, the spatial dimensions of the resulting feature map are reduced to 64×64, representing the resolution of the extracted features. We focus exclusively on 1-way 1-shot learning and perform 5-fold cross-validation across all experiments. Our LMS-Net is trained episodically for $35000$ iterations using the stochastic gradient descent optimizer \cite{Ref34} with a batch size of $1$. The initial learning rate is set to $1\times 10^{-3}$ and decays by a factor of $0.98$ every $1000$ iterations.
\subsection{Evaluation Metric}

\begin{figure}
    \centering
    \includegraphics[width=1\linewidth]{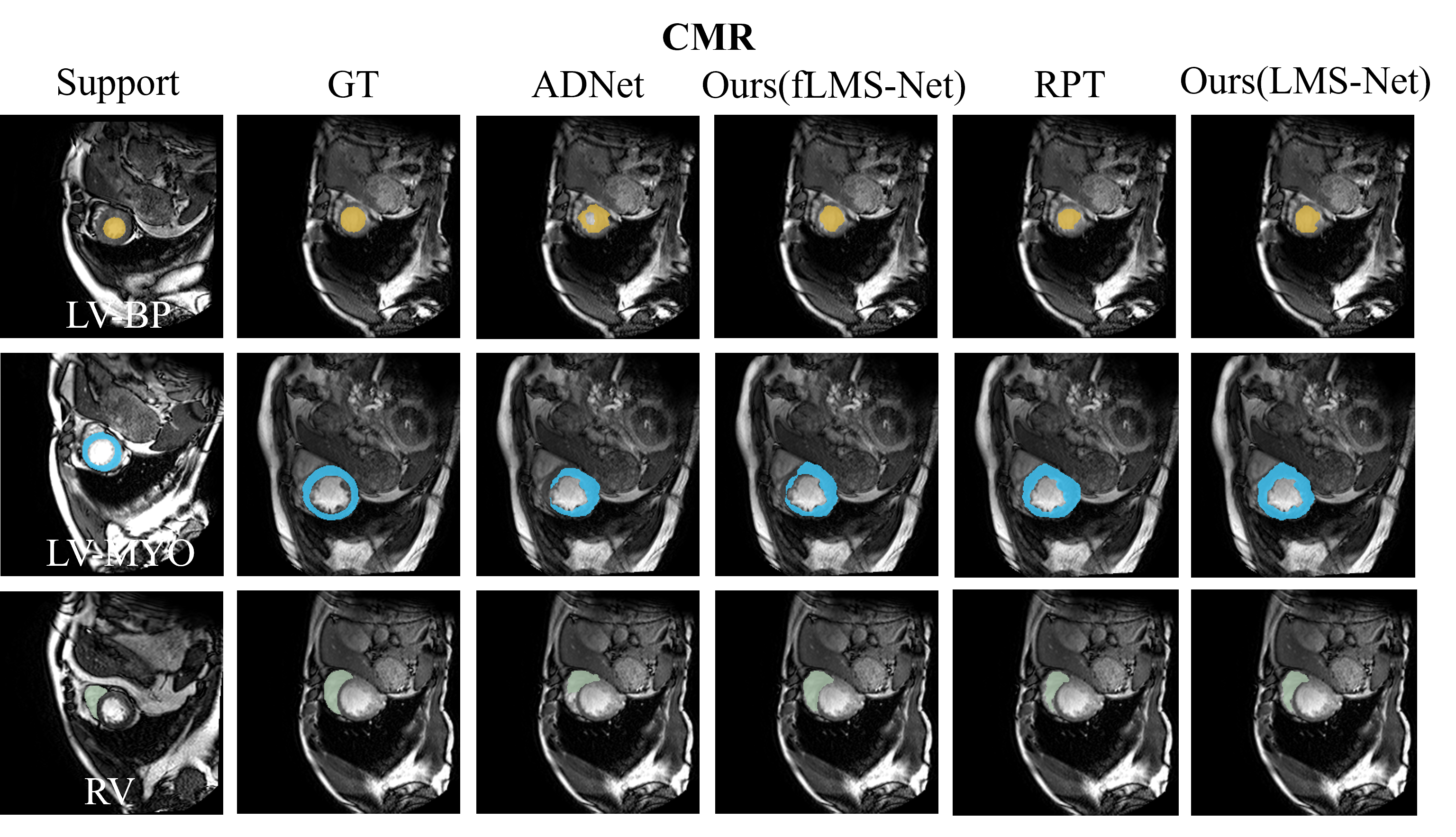}
    \caption{Qualitative comparison of our proposed LMS-Net with other medical FSS methods on the CMR dataset.}
    \label{CMR}
\end{figure}

Following established practices \cite{Ref28,Ref26}, the Dice Similarity Coefficient (DSC) is employed as the evaluation metric to assess the model performance. The DSC quantifies the overlap between two anatomical regions by calculating the similarity between the ground truth foreground mask $\hat{u}_1^q$ and the predicted foreground mask $u_1$. The coefficient is defined as:
\begin{equation}
\operatorname{DSC}(u_1, \hat{u}_1^q)=\frac{2|u_1 \cap \hat{u}_1^q|}{|u_1|+|\hat{u}_1^q|}.
\end{equation}
\subsection{Comparison with State-of-the-Art FSS Methods}
\subsubsection{Quantitative Evaluation}

We present a comprehensive summary of the quantitative results for the proposed LMS-Net across three general datasets in Table~\ref{tab:1}. Among methods that do not update prototypes, fLMS-Net consistently outperforms ADNet. In contrast, for prototype-updating methods, LMS-Net consistently surpasses RPT. Specifically, fLMS-Net improves the mean DSC by $1.00\%$, $2.12\%$, and $4.14\%$ over ADNet on Abd-CT, Abd-MRI, and CMR datasets, respectively. Similarly, LMS-Net achieves improvements over RPT, with gains of $3.00\%$, $0.28\%$, and $1.00\%$ on the same datasets. These results highlight the strong segmentation performance of the proposed interpretable networks, with fLMS-Net excelling in non-prototype updating methods and LMS-Net demonstrating consistent advancements in prototype-updating approaches.
\subsubsection{Qualitative Evaluation}
To intuitively demonstrate the effectiveness of LMS-Net and fLMS-Net, we visualize experimental results on Abd-CT, Abd-MRI, and CMR datasets, as presented in Fig. \ref{CT_MR} and Fig. \ref{CMR}. The results clearly show that the proposed LMS-Net achieves superior performance compared to other methods. Additionally, fLMS-Net outperforms ADNet, as the proposed approach explicitly incorporates the deep prior of the mask, leading to segmentation results that better preserve the structural integrity of semantic regions.

\begin{figure}
    \centering
    \includegraphics[width=1\linewidth]{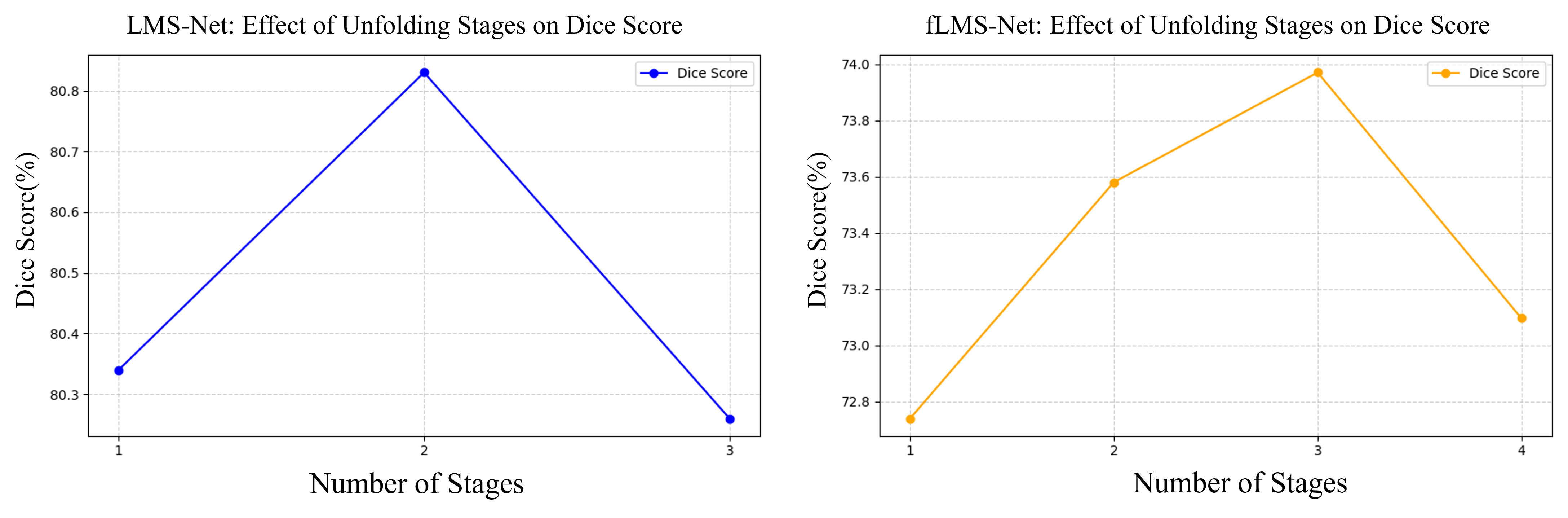}
    \caption{The Dice Score curve on Synapse-CT dataset with a
different number of stages K.}
    \label{fig:abl stage}
\end{figure}

\subsection{Ablation Analysis}
\subsubsection{Effect of the Number of Stages $K$}
To evaluate the impact of the number of iteration stages $K$ on segmentation performance, we compared the proposed LMS-Net and fLMS-Net across different stage settings. Fig. \ref{fig:abl stage} presents the mean Dice Score from 5-fold cross-validation on the Synapse-CT dataset. The results indicate that LMS-Net achieves optimal performance with two cascades, while fLMS-Net performs best with three cascades.
\subsubsection{Iteration results}
\begin{figure}
    \centering
    \includegraphics[width=1\linewidth]{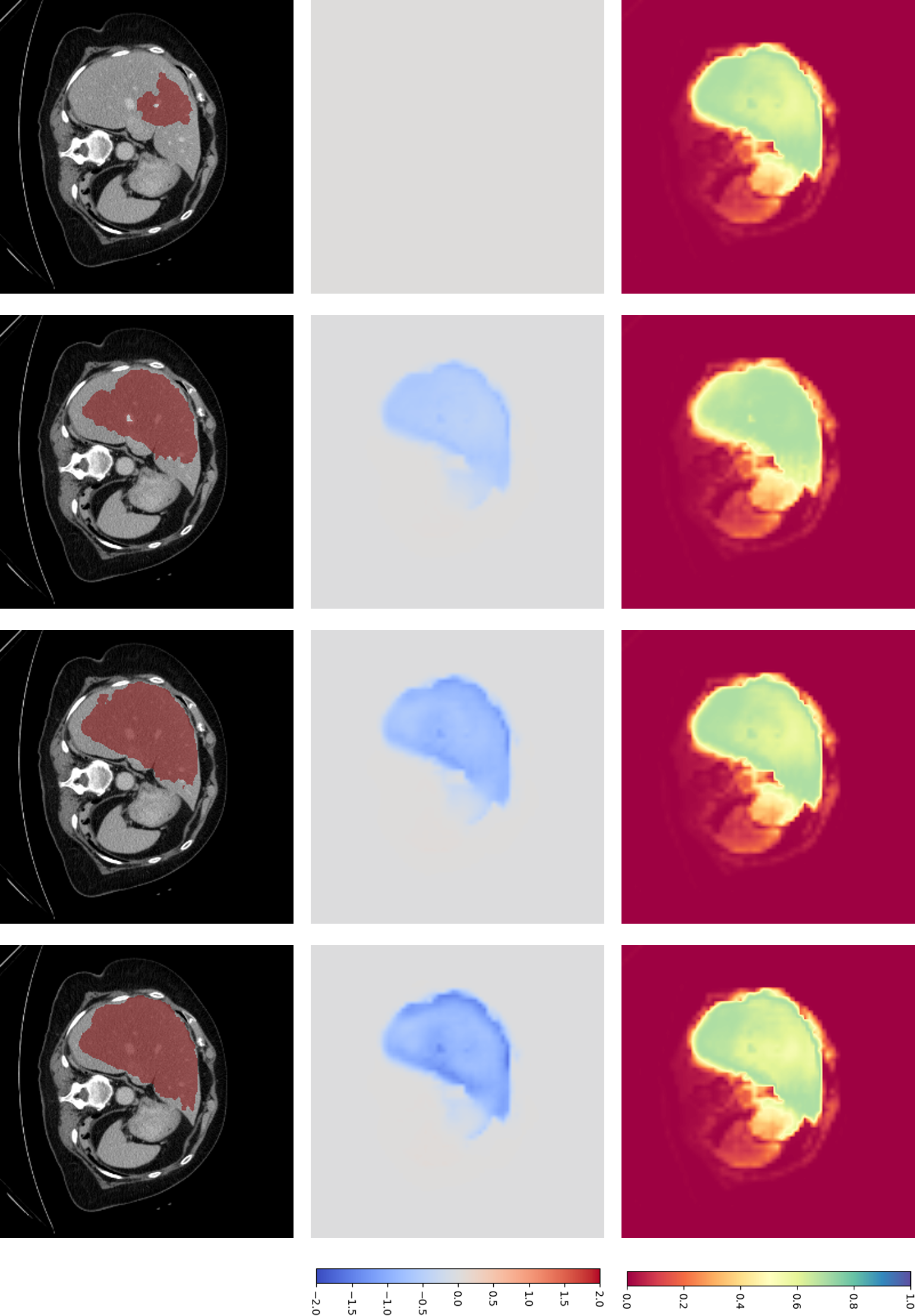}
    \caption{Iterates 0,1,2,3 in the fLMS-Net when refining the prediction mask. Left: predicted mask $u_1^k$ after binarization. Middle: predicted noise $v_1^k$. Right: Entropy of predicted mask $E(u^k)$.}
    \label{fig:iter}
\end{figure}

To better understand how the spatial prior of semantic regions is captured, we visualize the intermediate iterations of fLMS-Net. In Fig. \ref{fig:iter}, we show several iterations of $u_1^k$, $v_1^k$, and $E(u^k)$, where $E(u^k)$ highlights regions with higher uncertainty. Notably, we observe that the initial mask is suboptimal, and in the uncertain region, only a small fraction of the points is correctly classified. However, as the iterations progress, an increasing number of uncertain points are correctly classified. This behavior illustrates the balance between the data fidelity term and the prior term during training. If the network’s mask refinement capability is strong, even an imperfect initial mask can be improved, yielding an accurate final mask.

\subsubsection{Effect of key components}
\begin{figure*}
    \centering
    \includegraphics[width=1\linewidth]{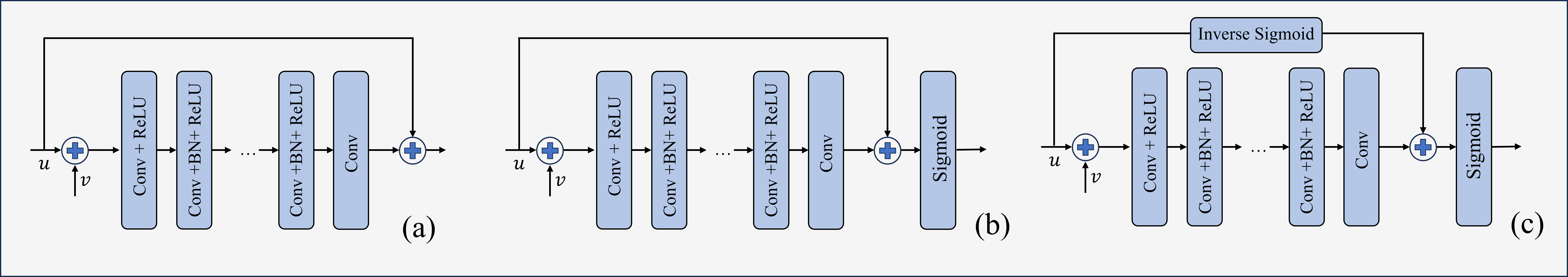}
    \caption{Schematic illustrations of the variants of Mask Denoiser, where (c) is our proposed Mask Denoiser.}
    \label{fig:variant}
\end{figure*}

\begin{table}[ht]
\caption{Quantitative results of the LMS-Net with and without PD-Net on Synapse-CT and  CHAOS-T2 datasets.}
\label{tab:2}
\centering

\begin{tabular}{c l l l l l l  }
\hline

Dataset & Method & LK & RK & Liver & Spleen & Mean  \\
\hline
\multirow{2}{*}{Abd-CT} & w/o PD-Net & 81.14 & 75.95 & 82.04 & 81.98 & 80.27  \\
                        & w PD-Net & 82.46 & 79.32 & 78.57 & 82.98 & \textbf{80.83 }  \\
\hline
\multirow{2}{*}{Abd-MRI} & w/o PD-Net & 79.31 & 89.39 & 83.9 & 73.19 & 81.45  \\
                        & w PD-Net & 82.85 & 89.34 & 83.71 & 74.97 & \textbf{82.72}   \\
\hline

\end{tabular}

\end{table}




In the ablation study for PD-Net, we evaluate the impact of replacing the PD-Net component with DC from the network. This variant is referred to as the proposed LMS-Net with DC (w/o PD-Net), which corresponds to the removal of the deep prior term in \eqref{deqn_ex11a}. The performance of these two networks is summarized in Table \ref{tab:2}. Specifically, PD-Net improves the Dice score by $0.56\%$ and $1.27\%$ on the Abd-CT and Abd-MRI datasets, respectively, highlighting the effectiveness of the proposed PD-Net.


\begin{table}[ht]
\caption{Ablation study on LMS-Net with the new variants of MD on Synapse-CT
dataset.}
\label{tab:3}
\centering

\begin{tabular}{c l l l l l  }
\hline
MD Variants & LK & RK & Liver & Spleen & Mean  \\
\hline
(a) & 65.78 & 58.37 & 74.05 & 68.69 & 66.72  \\
(b) & 78.28 & 74.08 & 83.12 & 81.90 & 79.35  \\
(c) & 82.46 & 79.32 & 78.57 & 82.98 & \textbf{80.83}   \\
\hline
\end{tabular}
\end{table}
\subsubsection{Analysis of MD}
Our MD, introduced in Section. \ref{Iter}, is specifically designed to learn the implicit prior of the predicted mask. Two key modifications distinguish it from the vanilla DnCNN: the inclusion of a Sigmoid operator in the final layer and an inverse Sigmoid to facilitate identity mapping. 
To validate these modifications, we conduct ablation experiments with two basic variants. The structures of the variants correspond to the illustrations in Fig. \ref{fig:variant}, where (c)
is our proposed MD. The results presented in Table \ref{tab:3} demonstrate that the proposed MD achieves the best performance, highlighting the necessity and effectiveness of these adjustments.
\section{Conclusion}
In this paper, we propose the LMS-Net for the FSS task. The network architecture is derived from the LMS model, which naturally combines the strengths of pixel-to-prototype comparison with the capabilities of deep priors. Our approach demonstrates that the primal-dual algorithm enables the mask update task to be decoupled into two subproblems: a simple data subproblem with a closed-form solution and a prior subproblem efficiently handled by a CNN denoiser. It can be seen as a natural extension of unfolding methods from the field of image reconstruction to image segmentation, using similar mathematical frameworks and optimization strategies. Extensive experiments and intermediate results validate the effectiveness and robustness of the proposed method.

The general idea behind the proposed approach is not confined to FSS applications. For instance, in broader semantic segmentation tasks, the LMS model can be unfolded to derive network architectures, with task-specific adaptations for prototype initialization. More importantly, in the context of medical imaging, this framework offers significant potential to improve both the interpretability and effectiveness of segmentation tasks, particularly in clinical settings where high-quality labeled data are scarce.

\end{document}